\documentclass{article}

\usepackage{microtype}
\usepackage{graphicx}
\usepackage{subfigure}
\usepackage{booktabs} % for professional tables

\usepackage{hyperref}

\usepackage[accepted]{icml2025}

\usepackage{amsmath}
\usepackage{amssymb}
\usepackage{mathtools}
\usepackage{amsthm}

\usepackage{amsfonts} % For mathematical fonts

\usepackage{graphicx} % To include graphics
\usepackage{enumitem} % For customizing lists

\usepackage{multirow}

\usepackage[capitalize,noabbrev]{cleveref}

\theoremstyle{plain}

\theoremstyle{definition}

\theoremstyle{remark}

\newcommand{\our}{{\textsc{BuildEvo}}}

\usepackage[textsize=tiny]{todonotes}

\icmltitlerunning{Submission and Formatting Instructions for ICML 2025}

\begin{document}

\twocolumn[
% \icmltitle{From Built Environment Physics to Predictive Power:\\ Informed Pre-processing for Deep Energy Models}
\icmltitle{\our: Designing Building Energy Consumption Forecasting Heuristics
via LLM-driven Evolution}



\icmlsetsymbol{equal}{*}

\begin{icmlauthorlist}
\icmlauthor{Subin Lin}{nus}
\icmlauthor{Chuanbo Hua}{kaist}
\end{icmlauthorlist}

\icmlaffiliation{nus}{Department of the Built Environment, National University of Singapore}
\icmlaffiliation{kaist}{Department of Industrial and Systems Engineering, Korea Advanced Institute of Science and Technology}

\icmlcorrespondingauthor{Chuanbo Hua}{cbhua@kaist.ac.kr}

\icmlkeywords{Building Energy, Large Language Models, Evolutionary Algorithms}

\vskip 0.3in
]



\printAffiliationsAndNotice{}  % leave blank if no need to mention equal contribution

\begin{abstract}
Accurate building energy forecasting is essential, yet traditional heuristics often lack precision, while advanced models can be opaque and struggle with generalization by neglecting physical principles. This paper introduces \our, a novel framework that uses Large Language Models (LLMs) to automatically design effective and interpretable energy prediction heuristics. Within an [evolutionary/iterative-refinement] process, \our guides LLMs to construct and enhance heuristics by systematically incorporating physical insights from building characteristics and operational data (e.g., from the Building Data Genome Project 2). Evaluations show \our achieves state-of-the-art performance on benchmarks, offering improved generalization and transparent prediction logic. This work advances the automated design of robust, physically grounded heuristics, promoting trustworthy models for complex energy systems.
\end{abstract}

\section{Introduction}
\label{sec:introduction}

Accurate building energy forecasting is a cornerstone of modern energy management, vital for enhancing energy efficiency, ensuring grid stability, and enabling effective building operations \citep{senthil2024enhancing}. Its widespread applications underpin demand-side response programs, the optimization of heating, ventilation, and air conditioning (HVAC), and the seamless integration of renewable energy sources \citep{chen2025systematic,mariano2020review}. Modeling the complex, interacting energy dynamics inherent in buildings—influenced by their physical structure, operational schedules, and environmental conditions—is thus a central challenge for smart grid balancing and optimized facility management \citep{bayasgalan2024comprehensive,thomas2020energy}.

However, achieving consistently accurate building energy forecasts is fundamentally difficult. Energy consumption patterns are notoriously complex, exhibiting non-linear behaviors and significant stochasticity due to fluctuating weather, diverse occupant activities, varying building thermal properties, and dynamic equipment states \citep{lim2017review}. Consequently, as evidenced by datasets like the Building Data Genome Project 2 (BDG2), even architecturally similar buildings can display vastly different energy profiles. Early attempts to capture these dynamics often relied on heuristic methods. These included rule-based systems derived from expert knowledge, simplified physical or resistance-capacitance (RC) models, and grey models (e.g., GM(1,1)) for trend estimation with limited data \citep{hu2020energy,pena2016rule,li2014review}. While offering interpretability, such handcrafted heuristics frequently suffer from limited accuracy and poor generalization, as their predefined rules and parameters are challenging to design and calibrate effectively for the diverse and dynamic nature of real-world energy consumption.

Deep learning (DL) methods have since emerged as powerful alternatives, with models like Long Short-Term Memory (LSTM) networks and Transformer-based architectures (e.g., Informer) demonstrating improved predictive accuracy \citep{Zhou2021Informer, khalil2022machine}. Despite their strengths, these DL approaches often present significant practical challenges. They can demand substantial computational resources and vast datasets for training, frequently operate as "black boxes" whose predictions lack transparency and interpretability \citep{runge2021review}, and may exhibit poor generalization to buildings or conditions not seen during training. This latter issue is particularly acute when standard data pre-processing overlooks or inadequately incorporates inherent physical knowledge of the built environment \citep{jiang2024modularized,von2024combining,miller2019more}. These limitations often necessitate a difficult trade-off with the clarity and robustness offered by simpler heuristic methods, motivating our research question: \emph{Can we automate the design of accurate and interpretable building energy forecasting heuristics that effectively leverage physical knowledge, thereby bridging the generalization and explainability gap with deep learning methods?}

Inspired by the burgeoning field of automated algorithm design that combines Large Language Models (LLMs) with Evolutionary Algorithms (EAs) \citep{romera2024mathematical,guo2024connecting}, we propose a novel approach to this problem. We hypothesize that the synergy between the generative and reasoning capabilities of LLMs and the structured search and optimization power of EAs can overcome traditional limitations of manual heuristic design, yielding sophisticated, physically-grounded energy forecasting heuristics. To this end, we introduce \our{}, a novel framework wherein LLMs operate within an evolutionary loop to iteratively generate, evaluate, and refine building energy forecasting heuristics directly from observational data, critically guided by underlying physical principles.

Our main contributions are as follows:
\begin{itemize}
    \item We present \our{}, a novel framework to integrate Large Language Models with Evolutionary Algorithms (EA) for the automated discovery and design of effective, explainable, and specifically physically-informed building energy consumption forecasting heuristics.
    \item We develop and incorporate mechanisms within \our{} that enable the systematic embedding of physical building information to the LLM-driven heuristic generation and refinement process, enhancing predictive performance and ensuring the physical plausibility of the generated heuristics.
    \item We conduct empirical validation of \our{} on public datasets (e.g., BDG2), demonstrating its capability to generate interpretable heuristics that achieve competitive forecasting accuracy and exhibit generalization capabilities across diverse building types and energy consumption patterns.
\end{itemize}

\section{Related Works}
\label{sec:related-works}

\paragraph{Heuristic Methods for Building Energy Consumption Forecasting}

Heuristic approaches provide simpler, often more interpretable alternatives for building energy forecasting, especially with limited data or when expert knowledge is paramount. These include rule-based systems and expert systems \citep{pena2016rule} that leverage predefined operational logic, grey models like GM(1,1) \citep{hu2020energy} for trend forecasting with sparse data, and simplified physical or grey-box models \citep{li2021grey} utilizing fundamental principles with reduced complexity. While typically less precise than deep learning, their transparency and lower computational cost make them useful for initial assessments or specific control applications.

\paragraph{Learning-based Methods for Building Energy Consumption Forecasting}

Machine learning and deep learning have notably advanced building energy consumption forecasting, yet challenges in handling complex time series data persist. Initial statistical models like Linear Regression (LR) and machine learning methods such as Support Vector Regression (SVR) and Random Forest (RF) provided foundational capabilities but often fell short with non-linear, non-stationary energy data. Deep learning, particularly Long Short-Term Memory (LSTM) networks\cite{ahmed2022review} and Transformer-based models like Informer \citep{Zhou2021Informer}, offered significant improvements for long-sequence forecasting. 
%%%%%%%%%%%%%%%%%%%%%%%%%%
More recently, hybrid strategies combining signal decomposition techniques like Ensemble EMD (EEMD) with deep learning predictors (e.g., EEMD-Informer) and hyperparameter optimization using methods like Particle Swarm Optimization (PSO) have become prominent. Despite these advancements, effectively managing data volatility, optimizing complex model parameters (as in PSO-Informer), and ensuring robust long-term prediction accuracy remain critical challenges, motivating our EEMD-PSO-Informer approach.

\section{Methodology}
\label{sec:methodology}

\begin{figure}[t]  
    \centering
\includegraphics[width=0.45\textwidth]{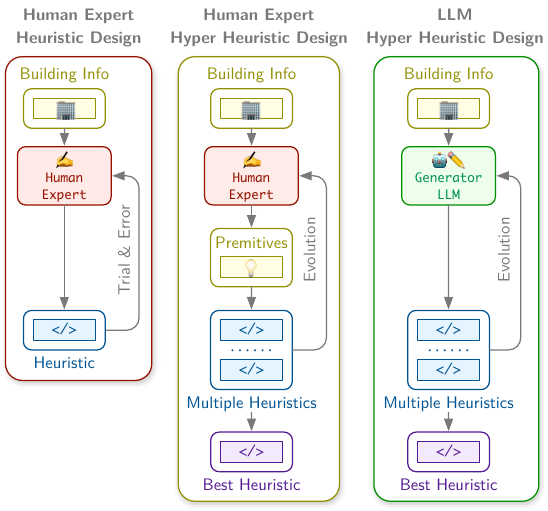}
    \caption{Comparision between human expert heuristic design, human expert hyper heuristic design, and LLM hyper heuristic design.}
    \vspace{-3mm}
    \label{fig:buildevo-overview}
\end{figure}

\subsection{Problem Definition}
\paragraph{Task}
We address building-level energy consumption forecasting using deep learning models.
The task is, for each building \({b}\) and associated energy meter \({m}\), to predict its future energy consumption sequence \(Y_{b,m} = (C_{b,m}^{T_{\mathrm{obs}}+1}, \dots, C_{b,m}^{T_{\mathrm{obs}}+T_{\mathrm{pred}}})\), where \(C_{b,m}^t \in \mathbb{R}\) is the consumption at future timestep \(t\).
This prediction is based on the observed history \(H_{b,m} = (\mathcal{S}_{b,m}^1, \dots, \mathcal{S}_{b,m}^{T_{\mathrm{obs}}})\), where \(\mathcal{S}_{b,m}^t = (C_{b,m}^t, X_b^t)\) comprises the past energy consumption \(C_{b,m}^t\) and relevant exogenous features \(X_b^t\) (including building characteristics, historical weather, and calendar effects) at timestep \(t\).
Additionally, known future exogenous inputs \(X_{b}^{\text{future}} = (X_b^{T_{\mathrm{obs}}+1}, \dots, X_b^{T_{\mathrm{obs}}+T_{\mathrm{pred}}})\) (e.g., weather forecasts, future calendar effects) are utilized.
Models are trained on a training set \(D_{\text{train}}\) and evaluated on a test set \(D_{\text{test}}\).

\paragraph{Metrics}
We evaluate the predictive performance of the deep learning models using standard regression metrics computed on the test set \(D_{\text{test}}\). These include, but are not limited to: Root Mean Squared Error (RMSE), Mean Absolute Error (MAE), and Mean Absolute Percentage Error (MAPE).
These metrics collectively provide insights into the accuracy (RMSE, MAE) and scale-independent relative error (MAPE) of the forecasts.

\subsection{Dataset and Preprocessing}
We utilized the Building Data Genome Project 2 (BDGP2) dataset \cite{miller2020building}, containing building metadata, weather, and multi-modal energy data, which was preprocessed to create a robust dataset for precise energy prediction. After integrating and temporally aligning these diverse sources, a hierarchical, domain-knowledge-driven imputation strategy addressed missing values across meter types. This involved rule-based imputation (leveraging metadata for operational logic), model-based imputation (using machine learning with temporal, weather, and building metadata features), supplemented by donor-based methods and minimal interpolation. The resulting complete and coherent dataset features intelligently imputed values reflecting physical and operational context, preserving data integrity (details in Appendix~\ref{sec:appendix}) and forming a solid foundation for subsequent prediction modeling.

% \subsection{Dataset and Preprocessing}
% We utilized the Building Data Genome Project 2 (BDGP2) dataset \cite{Miller2020-yc}, comprising building metadata, weather, and multi-modal energy data. Our preprocessing aimed to create a robust dataset for energy prediction by addressing missing data using built-environment-informed techniques. This involved integrating and temporally aligning (e.g., timestamp localization, weather resampling) these diverse data sources into a unified dataset per building. A hierarchical, domain-knowledge-driven imputation strategy addressed missing values across eight distinct meter types. Key steps included rule-based imputation (leveraging metadata like \texttt{primaryspaceusage} for operational logic) and model-based imputation using machine learning regressors with temporal, weather, and building metadata features. Donor-based methods from similar buildings and minimal temporal interpolation handled any residual gaps. The resulting complete and coherent dataset features intelligently imputed values reflecting physical and operational context, preserving data integrity (details in Appendix~\ref{sec:appendix}) and forming a solid foundation for robust energy prediction modeling.

\begin{table*}[!tbp] % Use table* for a two-column spanning table, [t] for top placement
\small
\caption{Performance results (MAPE (\%) / RMSE / MAE) of different prediction methods across generalization.}
\label{performance-table-generalization}
\vskip 0.15in
\begin{center}
\begin{small}
\begin{tabular}{l||c|c|c|c|c}
\toprule
\textbf{Method} & \parbox{1.8cm}{\centering To College Classroom} & \parbox{1.8cm}{\centering To Public Services-B} & \parbox{1.8cm}{\centering To College Dormitory} & \parbox{1.8cm}{\centering To Office Fox} & \parbox{1.8cm}{\centering To Public Services-R} \\
\midrule
LR & 12.99 / 9.76 / 6.76 & 16.45 / 9.02 / 6.23 & 84.11 / 4.95 / 2.97 & 21.78 / 8.11 / 4.71 & 65.08 / 8.36 / 5.51 \\
SVR & 9.71 / 8.20 / 5.12 & 9.36 / 6.00 / 3.63 & 84.09 / 4.96 / 2.86 & 14.55 / 5.71 / 3.54 & 62.36 / 8.13 / 5.51 \\
RF & 11.04 / 8.49 / 5.76 & 10.66 / 6.51 / 4.03 & 83.32 / 4.95 / 2.96 & 19.69 / 6.38 / 4.21 & 58.51 / 8.33 / 5.71 \\
LSTM & 10.00 / 8.09 / 5.23 & 9.83 / 6.42 / 3.73 & 79.62 / 4.88 / 2.86 & 14.04 / 5.28 / 3.43 & 61.61 / 8.25 / 5.65 \\
Informer & 8.69 / 6.88 / 4.61 & 8.64 / 5.84 / 3.27 & 29.79 / 3.84 / 2.57 & 10.84 / 4.60 / 2.88 & 47.55 / 6.47 / 4.32 \\
PSO-Informer & 8.01 / 6.64 / 4.28 & 8.58 / 5.78 / 3.31 & 27.15 / 3.75 / 2.50 & 10.72 / 4.53 / 2.84 & 46.60 / 6.35 / 4.14 \\
\midrule
- PIFL & 9.10 / 7.35 / 4.75 & 8.90 / 5.80 / 3.38 & 73.50 / 4.42 / 2.60 & 12.75 / 4.80 / 3.10 & 57.00 / 7.50 / 5.12 \\
\our & \textbf{7.10 / 5.90 / 3.80} & \textbf{7.60 / 5.15 / 2.95} & \textbf{24.00 / 3.35 / 2.22} & \textbf{9.50 / 3.79 / 2.52} & \textbf{43.03 / 5.50 / 3.60} \\
\bottomrule
\end{tabular}
\end{small}
\end{center}
\vskip -0.1in
\label{tab:main-results}
\end{table*}

\subsection{Evolutionary Framework}
\label{subsec:evolutionary-framework}

Our evolutionary framework adapts the Reflective Evolution approach \citep{ye2024reevo}, employing Large Language Models (LLMs) for core genetic operators (initialization, crossover, mutation), guided by reflective analyses of energy forecasting heuristic performance.

\paragraph{Initial Population}
The process starts by providing the generator LLM with a task specification for building energy forecasting—including inputs (e.g., historical energy, weather, building metadata like \texttt{sqft}, \texttt{primaryspaceusage}), outputs (future energy consumption), and the objective $J$ (e.g., minimizing RMSE)—alongside a basic energy heuristic example (e.g., a persistence model). The LLM then generates an initial population of $N$ diverse code-based energy forecasting heuristics.

\paragraph{Selection for Crossover}
Parent heuristics for crossover are chosen from successfully executed candidates, balancing exploration (70\% random selection) and exploitation (30\% from elite performers with the lowest forecasting error $J$).

\paragraph{Reflections}
\our{} utilizes two reflection types \citep{ye2024reevo}. \textit{Short-term reflections} compare selected parent heuristics to guide crossover, for instance, by identifying better use of weather data. \textit{Long-term reflections} accumulate insights across generations, pinpointing effective design patterns for energy heuristics, especially concerning the integration of physical building parameters, and generating textual "verbal gradients" \citep{pryzant2023automatic} for the LLM.

\paragraph{Crossover}
Guided by short-term reflections comparing parent performance, the LLM combines elements (e.g., code segments, rules) from two parent heuristics. This aims to synthesize offspring with potentially superior energy forecasting logic, such as improved handling of building metadata or weather impacts.

\paragraph{Elitist Mutation}
The LLM mutates an elite heuristic (potentially selected by CGES, see below), informed by long-term reflections. This focuses on refining strategies, particularly for incorporating physical insights like occupancy schedules inferred from \texttt{primaryspaceusage} or weather normalization techniques.

\subsection{Cross-Generation Elite Sampling for Energy Heuristics}
\label{subsec:cross-generation-elite-sampling-energy}

Standard evolutionary search for complex building energy forecasting heuristics can stagnate in local optima \citep{osuna2018runtime}, as simple mutations often yield only minor improvements. To bolster exploration, \our{} adapts Cross-Generation Elite Sampling (CGES). CGES maintains a historical archive of high-performing energy forecasting heuristics from all past generations. For mutation, instead of only using the current best, an elite heuristic is sampled from this archive via a Softmax distribution based on past forecasting errors $J$. Reintroducing and modifying these historically successful, diverse strategies (e.g., heuristics adept for specific building types or weather conditions) enhances exploration, aiding escape from local optima and fostering the discovery of more robust energy heuristics.

\subsection{Physical Insights Feedback Loop}
\label{subsec:physical-insights-feedback-loop}

The aggregate forecasting error $J$ does not reveal the efficacy of specific internal components or rules within a heuristic, especially those leveraging physical building knowledge. \our{} incorporates a Physical Insights Feedback Loop (PIFL) to address this. This loop analyzes the contribution of distinct internal logical segments or physically-informed rules—such as those for calculating base load via \texttt{sqft} and \texttt{primaryspaceusage}, applying weather-dependent adjustments, or modeling occupancy-driven variations. Statistical feedback on the empirical utility of these components is provided to the reflector and mutation LLMs. This guides them to refine how the heuristic effectively utilizes physical building insights and operational contexts for improved performance, steering the evolution towards more physically grounded and accurate energy forecasting heuristics.

\section{Experiments}
\label{sec:result}

\subsection{Experimental Setup}

\paragraph{Baselines}
We compare the heuristics generated by \our{} against several benchmark methods commonly used or relevant for building energy consumption forecasting. These include statistical and machine learning models such as Linear Regression (LR) \citep{bishop2006pattern}, Support Vector Regression (SVR), and Random Forest (RF); deep learning models like the Long Short-Term Memory network (LSTM) \citep{hochreiter1997long} and Informer \citep{Zhou2021Informer}, the latter being specifically designed for long-term series forecasting with a self-attention mechanism and an encoder-decoder architecture; and optimization-enhanced models including Particle Swarm Optimization (PSO) \citep{kennedy1995particle}, a classical population-based stochastic optimization technique, and PSO-Informer, which utilizes PSO to optimize Informer's parameters. The performance of our proposed method is evaluated against these established approaches.

\paragraph{Hardware and Software}
All experiments were conducted on a workstation equipped with an AMD Ryzen 9 7950X 16-Core Processor and a single NVIDIA RTX 5090 GPU. The \our{} framework generates building energy forecasting heuristics as executable Python code snippets in a Python 3.12 environment, employing Google's Gemini 2.0 Flash model \citep{GoogleGeminiFlash2.0}.

\subsection{Main Results}

We report the performance results against the benchmark methods in \cref{tab:main-results}. The heuristics generated by \our{} demonstrate competitive performance. This establishes \our{} as a promising approach for generating effective energy forecasting heuristics. Note that with \our, the algorithm doesn't have learnable parameters compared with previous neural baselines.  

We also do the ablation study about the PIFL as present in \cref{tab:main-results}. Without the PIFL, \our still achieves a competitive performance compared to the neural baselines. By adding the PIFL, the performance of \our has significantly improved, which validates the effectiveness of the PIFL.

\subsection{Explainability}
A key advantage of \our{} is the generation of explainable Python code for its heuristics. Unlike the "black-box" nature of many DL models, the logic of the evolved heuristics can be inspected and understood. For instance, an evolved heuristic might explicitly combine rules based on time-of-day, weather inputs, and building metadata, offering transparent insights into its forecasting strategy. This interpretability is highly beneficial for building trust, debugging, and deploying in critical energy management systems.

\section{Conclusion}
\label{sec:conclusion}
We introduced \our{}, a novel framework leveraging LLM and EAs to automatically design physically-informed building energy forecasting heuristics. \our{} generates interpretable heuristics that achieve competitive accuracy and generalize well, as demonstrated on datasets like BDG2. This work offers a new path towards bridging the gap between complex "black-box" models and traditional heuristics, providing a method for creating transparent and effective forecasting tools.


\bibliography{ref}
\bibliographystyle{icml2025}

%%%%%%%%%%%%%%%%%%%%%%%%%%%%%%%%%%%%%%%%%%%%%%%%%%%%%%%%%%%%%%%%%%%%%%%%%%%%%%%
%%%%%%%%%%%%%%%%%%%%%%%%%%%%%%%%%%%%%%%%%%%%%%%%%%%%%%%%%%%%%%%%%%%%%%%%%%%%%%%
% APPENDIX
%%%%%%%%%%%%%%%%%%%%%%%%%%%%%%%%%%%%%%%%%%%%%%%%%%%%%%%%%%%%%%%%%%%%%%%%%%%%%%%
%%%%%%%%%%%%%%%%%%%%%%%%%%%%%%%%%%%%%%%%%%%%%%%%%%%%%%%%%%%%%%%%%%%%%%%%%%%%%%%
\newpage
\appendix
\onecolumn
\section{Appendix}
\label{sec:appendix}
\subsection*{Data Preprocessing and Loading Protocol}
Our study utilizes the Building Data Genome Project 2 (BDGP2) dataset \cite{miller2020building}, a comprehensive collection comprising building metadata, high-frequency weather data, and multi-modal energy consumption readings (electricity, chilled water, hot water, steam, gas, solar, irrigation, and water) across a diverse set of buildings. The primary objective of our data preprocessing pipeline is to create a robust, high-quality dataset suitable for precise energy usage prediction by addressing missing data through techniques informed by built environment characteristics.

\subsection{Data Integration and Temporal Alignment}

The initial phase involved meticulous integration of the disparate data sources. Building metadata, including structural details (e.g., \texttt{sqft}, \texttt{yearbuilt}), operational characteristics (e.g., \texttt{primaryspaceusage}), and geographical information (e.g., \texttt{timezone}, \texttt{site\_id}), was centralized. Meter readings from various cleaned CSV files, each representing a distinct energy type, were systematically processed. Weather data, including variables such as temperature, humidity, and solar irradiance, was also incorporated.

A critical step was ensuring temporal consistency across all datasets. Timestamps from meter and weather data were parsed and localized using timezone information from the metadata. Weather data, often recorded at a different frequency than meter readings, was resampled (e.g., to hourly averages or sums, as appropriate for the variable) to align with the temporal resolution of the energy consumption data. These datasets were then merged using \texttt{building\_id} (for metadata) and a combination of \texttt{site\_id} and the aligned \texttt{timestamp} (for weather data), creating a unified analytical dataset for each building and meter type.

\subsection{Missing Data Characterization}

Prior to imputation, a thorough analysis was conducted to characterize the nature and extent of missing values within each energy meter stream. This involved quantifying missing data percentages and identifying patterns, such as correlations with specific time periods (e.g., time of day, day of week, season), building archetypes (derived from \texttt{primaryspaceusage}), or particular weather conditions. This step was crucial for selecting appropriate imputation strategies.

\subsection{Imputation Strategy based on Built Environment Knowledge}

We employed a hierarchical imputation strategy, prioritizing methods that leverage domain-specific knowledge of building energy systems and operational characteristics over purely statistical approaches.

\subsubsection{Rule-Based and Knowledge-Driven Imputation}
Initial imputation was performed using predefined rules derived from building metadata and typical operational schedules. For instance, energy consumption for certain meter types (e.g., solar generation during nighttime, or HVAC in unoccupied commercial spaces during holidays inferred from \texttt{primaryspaceusage} and temporal features) was imputed with zero or near-zero values where logical. The presence or absence of specific utilities for a building, as indicated in the metadata, also guided this step.

\subsubsection{Model-Based Imputation}
For remaining gaps, machine learning models were developed to predict missing energy consumption values. For each meter type, regression models (e.g., Gradient Boosting Regressors or Random Forests) were trained on segments of data with complete records. Features for these models included:
\begin{itemize}
    \item \textbf{Temporal Features:} Cyclical time features (hour of day, day of week, month), holiday indicators, and trend components.
    \item \textbf{Weather Features:} Current and lagged weather variables (e.g., temperature, humidity, solar irradiance, wind speed), and derived features like heating/cooling degree days.
    \item \textbf{Metadata Features:} Building characteristics such as \texttt{sqft}, \texttt{primaryspaceusage} (typically one-hot encoded), and building age.
\end{itemize}
The trained models were then used to predict and fill missing consumption values, thereby incorporating the dynamic interplay between energy use, external conditions, and building properties.

\subsubsection{Donor-Based Imputation}
In instances where model-based imputation was challenging due to extended missing periods or unique building behaviors not well captured by general models, a donor-based approach was considered. This involved identifying "similar" buildings (based on \texttt{primaryspaceusage}, \texttt{sqft}, \texttt{timezone}, and other relevant metadata) and using their normalized consumption patterns during comparable temporal and weather contexts to inform imputation for the target building.

\subsubsection{Temporal Interpolation}
As a final step, for any very short, isolated missing data points (e.g., single-record gaps) not addressed by the above methods, standard temporal interpolation techniques (e.g., linear interpolation) were applied sparingly.

This multi-faceted imputation process was applied iteratively to each of the eight distinct meter types. The specific logic for model features and rule-based imputations was tailored where necessary to the unique physical drivers of each energy stream (e.g., solar irradiance for solar generation, temperature differentials for HVAC-related meters like chilled water or steam).

\subsection{Final Dataset Assembly}
\begin{figure}[ht]
\vskip 0.2in
\begin{center}
\centerline{\includegraphics[width=\columnwidth]{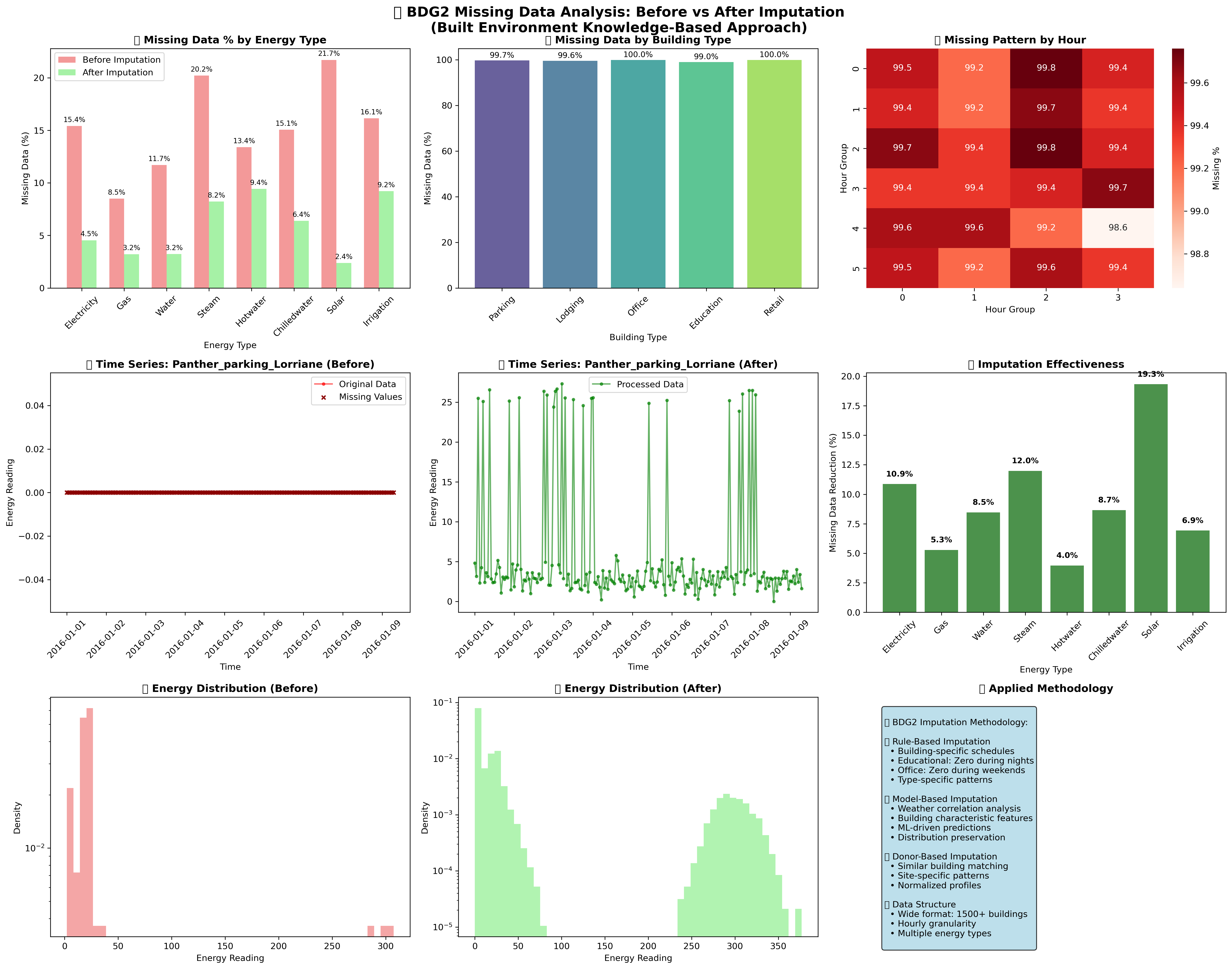}}
\caption{Visual overview of the built-environment-informed imputation process, demonstrating its effectiveness in reducing missing data while preserving key data characteristics across multiple dimensions.}
\label{fig:comparison}
\end{center}
\vskip -0.2in
\end{figure}

Upon completion of the imputation process for all meter types, the individually processed datasets were merged into a single, comprehensive time-series dataset. An overview of the imputation impact is presented in Figure \ref{fig:comparison}, which showcases the substantial reduction in missing data and the preservation of data integrity (e.g., temporal trends, statistical distributions) achieved through the built-environment-informed methodology. This final dataset, characterized by its completeness and the informed nature of its imputed values, serves as the foundation for subsequent machine learning model development for energy usage prediction.

\end{document}